\documentclass[letterpaper, 10 pt, conference]{ieeeconf}  

\IEEEoverridecommandlockouts                              

\usepackage[tracking=true]{microtype}
\usepackage{amsmath} 
\usepackage{amssymb}  
    \usepackage{graphicx}
\usepackage{epsfig} 
\usepackage{times} 
\usepackage{amsmath} 
\usepackage{amssymb}  
\usepackage{xspace}
\usepackage{lipsum} 
\usepackage{xcolor, soul}
\usepackage[table]{xcolor}
\usepackage{algorithm}
\usepackage{algpseudocode}
\usepackage{wrapfig}

\usepackage{minted}
\usepackage{float}
\usepackage{afterpage}
\setminted{fontsize=\footnotesize}
\usepackage{url}
\newcommand{\ve}[1]{\mathbf{#1}} 
\usepackage[dvipsnames]{xcolor}
\usepackage[font={footnotesize}]{caption}
\usepackage{booktabs} 
\usepackage{tikz}
\usetikzlibrary{positioning,shapes.geometric,arrows,fit,shapes.symbols,svg.path,tikzmark,calc}
\usepackage{textcomp}
\usepackage{listings}
\usepackage{subcaption}
\pgfdeclarelayer{background}
\pgfdeclarelayer{foreground}
\pgfsetlayers{background,main,foreground}
\usepackage{siunitx}
\usepackage{adjustbox}
\usepackage{array}
\usepackage{multirow}
\usepackage{hyperref}
\usepackage{siunitx}
\usepackage{gensymb}

\let\labelindent\relax 

\usepackage[inline]{enumitem} 

\usepackage[backend=biber,
            url=false,
            isbn=false,
            doi=false,
            backref=false,
            style=ieee,
            natbib=true,
            mincitenames=1,
            maxcitenames=1,
            citestyle=numeric-comp,
            sorting=none,
            block=none]{biblatex}
\renewcommand{\bibfont}{\small}
\newcommand{\algabbr}{TRACE\xspace}

\newenvironment{tightquote}
  {\list{}{\setlength{\leftmargin}{1em}%
           \setlength{\rightmargin}{1em}%
           \setlength{\topsep}{2pt}%
           \setlength{\parsep}{0pt}}%
   \item\relax}
  {\endlist}

\title{\LARGE \bf TRACE:\\Interactive Bi-Directional Tracing of Monochrome Cables Amid Clutter}

\author{Nidhya Shivakumar$^{*1}$, Ethan Ransing$^{*1}$, Josh Zhang$^{1}$, Shamak Gowda$^{1}$, Kevin Yang$^{1}$, \\ Miles Hua$^{1}$, Anika Agrawal$^{1}$, Justin Yu$^{1}$, Ken Goldberg$^{1}$ 
\thanks{$^{1}$ AUTOLab at the University of California, Berkeley}
\thanks{$^{*}$ Equal contribution}}

\begin{document}
\makeatletter
\newif\ifTitleFigDone
\TitleFigDonefalse

\let\@oldmaketitle\@maketitle
\renewcommand{\@maketitle}{%
  \@oldmaketitle
  \ifTitleFigDone\else
    \global\TitleFigDonetrue
    \begin{center}
      \begin{minipage}{\linewidth}
        \includegraphics[width=\linewidth]{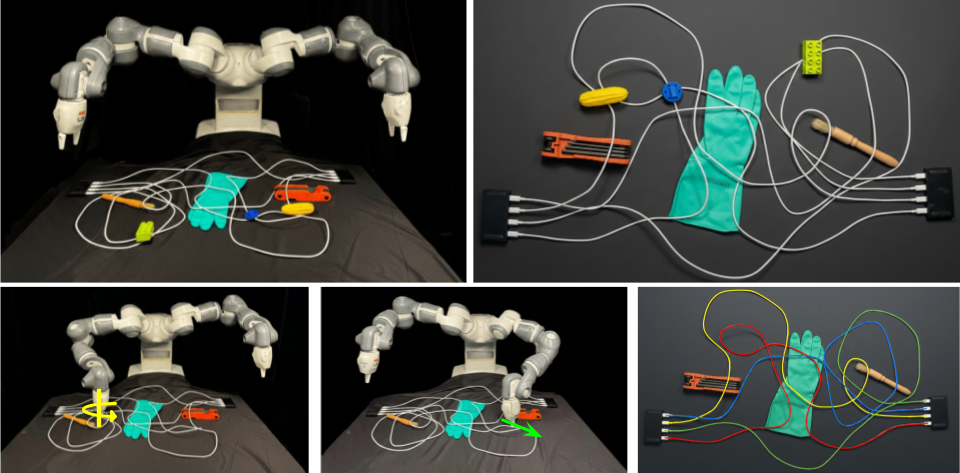}
        \captionof{figure}{(Top left): Initial scene with 4 monochrome cables, 3 foreground clutter objects, and 3 background clutter objects. (Top right): Overhead image of the initial scene as input to \algabbr. The novel interactive perception primitives used to resolve visual ambiguities are (Bottom left): Cluster Dilation and (bottom center): Divergence Push. (Bottom right) Color overlay confirms 100\% cables traced correctly with 3 foreground clutter objects.}
        \label{fig:concept}
        \vspace{-0.35cm}
      \end{minipage}
    \end{center}
  \fi
}
\makeatother
\maketitle
\thispagestyle{empty}
\pagestyle{empty}


\begin{abstract}

Accurate state estimation (tracing) of Deformable Linear Objects (DLOs) such as cables is a critical challenge for data centers, manufacturing, construction, homes, and surgery, where precise cable management directly impacts operational safety and efficiency. However, resolving the state of multiple monochrome cables amid foreground and background clutter poses challenges due to occlusions, overlap, and ambiguous crossings. We present Two-way Routing And Cable Estimation (\algabbr), which combines bi-directional cable tracing with interactive perception primitives—Divergence Push and Cluster Dilation—to actively resolve ambiguities. Evaluation with 110 physical experiments suggests that \algabbr can increase the percentage of cable length correctly traced in complex scenarios (with up to 4 cables and 40 crossings) from $\sim$60\% with the strongest prior method, HANDLOOM 2.0, to $\sim$90\%, outperforming RT-DLO, Nano Banana Pro, and ChatGPT 5.2 as well. For a trial run on a workstation with an NVIDIA GeForce RTX 4090 GPU, the average computation time is 0.4 seconds per cable. Project website: \url{https://trace-paper.github.io/}.

\end{abstract}

\begin{figure*}[h]
\centering
\vspace{0.5cm}
\includegraphics[width=\linewidth]{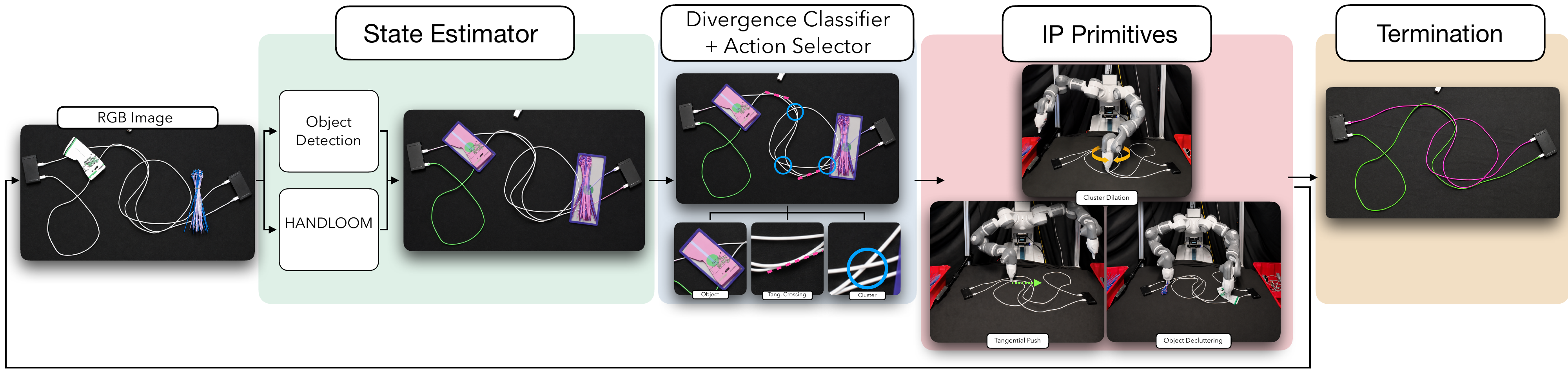} 
\caption{\small Overview of the \algabbr architecture using the MANIP \cite{yu2024manip} framework. The system first processes the overhead RGB image to detect occlusions by comparing the \algabbr-predicted cable state with object masks. If an occlusion is detected, a foreground object decluttering primitive is executed. Then, divergence points are identified via bi-directional tracing and classified as tangential crossings or cable clusters, activating the corresponding interactive perception primitive.}
\label{fig:sketch}
\vspace{-0.5cm}
\end{figure*}

\raggedbottom
\section{Introduction}

The manipulation and perception of Deformable Linear Objects (DLOs), such as cables, pose significant challenges in applications such as data center setup and maintenance, electrical wiring, product assembly, surgery, and construction. In many facilities, cluttered and tangled cables can reduce reliability, compromise safety, and complicate troubleshooting efforts. Unlike rigid objects, cables exhibit complex deformations, self-occlusions, and entanglements, which make them difficult to trace.

In this work, we focus on the particularly challenging task of monochrome cable configurations, in which individual cables are visually indistinguishable. For multi-colored cables, color segmentation can facilitate tracing by using chromatic contrast between two cables. Monochrome cable tracing algorithms must instead use geometric continuity and interactive disambiguation. Our only visual assumption is high contrast between the cables and the background; the cables are monochrome relative to one another, not necessarily white.

Single-cable tracing methods have shown promising results in controlled environments, using both analytical models based on geometric structures (V. Viswanath \textit{et al.} \cite{Shivakumar-RSS-22}) and data-driven techniques that trace cables from visual input (K. Shivakumar \textit{et al.} \cite{shivakumar2023sgtm}). Among these, Heterogeneous Autoregressive Learned Deformable
Linear Object Observation and Manipulation (HANDLOOM 1.0) \cite{viswanath2023handloom} is an RGB vision-based, autoregressive framework that sequentially traces cables by predicting the most likely cable path using learned visual features and contextual cues. While HANDLOOM 1.0 successfully traces individual cables in most structured environments, its extension to multi-cable scenarios faces significant challenges. We also evaluate Nano Banana Pro and ChatGPT 5.2, two general-purpose vision-language baselines, which fail to trace cables in dense monochrome settings (see Section V.F).

\textit{Interactive Perception} describes a class of techniques that allow robots to actively manipulate the environment to minimize state uncertainty and enhance observability (J. Bohg \textit{et al.} \cite{bohg2017interactive}). By physically interacting with object clutter and cables, robots can reveal occluded cable segments and resolve ambiguities at cluttered crossings. The Modular Architecture for Integrating Interactive Perception (MANIP) framework (J. Yu \textit{et al.} \cite{yu2024manip}) demonstrated that modular system architectures incorporating learned and model-based manipulation primitives can significantly improve single-cable estimation in the presence of multiple cables. Using the MANIP framework, HANDLOOM 2.0 applied interactive perception to improve tracing success rates by up to 88\%, but only traced 60\% to 68\% of cable length correctly in configurations with 3 or 4 cables.




In this work, we introduce \algabbr, a framework that extends previous multi-cable tracing approaches by introducing bi-directional tracing and new interactive perception primitives to resolve ambiguities caused by occlusions, crossings, and high-density cable arrangements. Unlike HANDLOOM 2.0, which relies on predefined heuristics (entropy, local cable density, trace length, and cable connector locations) to trigger intervention actions, \algabbr globally analyzes cable connectivity to identify regions of uncertainty and enforce consistency across multiple traces. \algabbr's cable disambiguation capabilities are shown in Figure~\ref{fig:concept}, and its architecture is visualized in Figure~\ref{fig:sketch}.

\noindent This paper makes the following contributions:
\begin{enumerate}
    \item The Two-way Routing And Cable Estimation (\algabbr) pipeline, a novel cable tracing algorithm for multiple monochrome cables in cluttered environments.
    \item A multi-step bi-directional tracing approach that identifies uncertainty and gaps in connectivity. 
    \item Two novel task-oriented robot interactive perception primitives (Divergence Push and Cluster Dilation) that combine geometric information and Hessian-based disambiguation to actively resolve crossings and occlusions.
    \item Physical evaluations on 110 unique real-world cluttered cable scenarios that suggest significant performance improvements over HANDLOOM 2.0, RT-DLO, Nano Banana Pro, and ChatGPT 5.2 \cite{yu2024manip}, \cite{RTDLO}.
\end{enumerate}
\section{Related Work}

Early vision-only single-cable tracing methods enforced spline smoothness  \cite{keipour2022deformable, kicki2023dloftbs, huang, choi2023mbest} or registered visual features across frames (\cite{Lui}, \cite{tang2017state}). While effective for isolated, well-lit cables, these analytical pipelines degrade when multiple cables overlap or self-intersect, as they lack a mechanism for resolving correspondence ambiguity in cluttered multi-cable scenes. Extensions to multiple cables and self-intersecting cables (\cite{xiang23multidlo, xiang23trackdlo}) improve robustness but remain limited under visual ambiguity.  

More recent learning-based approaches use instance segmentation and 3D reconstruction to trace DLOs. FastDLO \cite{caporali2022fastdlo} and RT-DLO \cite{RTDLO} perform real-time detection and extraction of the topology from semantic masks, but assume multi-colored cables. Although effective in isolating cable masks, these methods focus on per-frame segmentation and do not explicitly enforce cable connectivity consistency. Similarly, Zhaole \textit{et al.} \cite{zhaole23robust} introduce a 3D perception pipeline that recovers structure from segmented observations but does not handle densely cluttered configurations of visually indistinguishable cables. 

Other works focus on single DLO state estimation. M. Yan \textit{et al.} \cite{yan2020self} use a coarse-to-fine iterative refinement strategy by progressively expanding the region of interest. Likewise, K. Lv \textit{et al.} \cite{lv23occluded} learn a two-branch PointNet++ network to estimate the 3-D state of a single segmented deformable linear object behind foreground occlusions. Although effective for single-object manipulation, these methods do not extend to multi-cable tracing. HANDLOOM 1.0 \cite{viswanath2023handloom} uses a probabilistic learning-based approach that autoregressively predicts trace points using a UNet convolutional network applied to locally cropped, reoriented overhead images. Despite its effectiveness in structured environments, it degrades in cluttered multi-cable scenes where occluding objects and high cable density lead to failures. 


To improve tracing performance in ambiguous scenarios, interactive perception has been explored to resolve occlusions and crossings \cite{bohg2017interactive}. For instance, SGTM 2.0 \cite{shivakumar2023sgtm} combined an analytical cable tracing method with interactive manipulation primitives, demonstrating that targeted interactions significantly improve untangling. Building on these insights, HANDLOOM 2.0 \cite{yu2024manip} extends interactive perception to the cable state estimation task by integrating HANDLOOM 1.0 with manipulation policies for intervention. However, determining where and when to intervene via interactive perception in cable disambiguation scenes remains an open challenge, particularly as the complexity of multi-cable configurations increases. Unlike prior approaches, \algabbr explicitly targets multi-cable tracing under visual ambiguity by using bi-directional tracing, local cable geometry, interactive perception, and individual monocular RGB images; it does not rely on depth data to disambiguate occlusions.
\algabbr scales to complex cable configurations with up to 5 tangential crossings and 40 total crossings per scene, high-density clusters in which cables cover up to 90\% of a $\sim$6.45 cm$^2$ local region centered around a tangential crossing, within a $\sim$5500 cm$^2$ workspace, and up to 4 foreground objects cluttering a scene. We demonstrate transferable improvement in trace performance from a baseline evaluation against HANDLOOM 2.0 and RT-DLO.

\section{Problem Formulation}

\noindent We make the following assumptions:
\begin{enumerate}
    \item Monochrome cables can be visually distinguished from the background.
    \item All cable endpoints (i.e., connectors) are inserted into visible USB hubs and all cables are approximately coplanar.
    \item Monocular RGB images: no depth maps, stereo baselines, or calibrated range data. While RGB-D sensors have produced sparse and unreliable measurements for thin, deformable objects like cables in the past, recent RGB depth-estimation methods show promise; we leave this to future work.
\end{enumerate}

The workspace is defined as a planar region $\mathcal{W} \subset \mathbb{R}^2$, containing $n$ monochrome cables. Initially unknown, each cable $i$ is represented in the xy-plane by a continuous centerline function ${\theta_{i,t}(s)}$, which gives the true $(x,y)$ position of cable $i$ at timestep $t \in {\{0, 1, 2, ..., T_{max}\}}$ at normalized arc-length parameter $s \in [0,1]$ from its first connector $e$. The complete estimated cable state for all cables at timestep $t$ is given by ${\hat{\Theta}_t} = \bigcup_{i=1}^{n} {\hat\theta_{i,t}(s)}$, where ${\hat\theta_{i,t}(s)}$ is an estimate of ${\theta_{i,t}(s)}$. At each timestep $t$, the system receives a monocular RGB image observation $\ve{I_t}$ from a single overhead camera. The objective is to accurately reconstruct the complete ground truth cable state ${\Theta}_t$ and identify the corresponding connectors of each cable $i$. Connector locations are not provided; TRACE detects them from the RGB image.




In cases where visual ambiguities, occlusions, or dense cable configurations make passive observation insufficient, \algabbr uses robotic interactive perception primitives to actively manipulate the workspace. At each timestep $t$, the robot applies an interactive perception primitive $\ve{a_t} \in \mathcal{A}$, where $\mathcal{A}$ is the set of implemented primitives, producing a modified true cable state  ${\Theta}_{t+1}$ and updated estimate $\hat{\Theta}_{t+1}$.
\section{Method}

\subsection{Connector Detection}

To initialize the tracing algorithm, the system first identifies (but does not attempt to match) all cable connectors to the connector hubs as shown in Figure 1. We collected a dataset of labeled images containing random connector configurations to supervise a Faster R-CNN model \cite{ren2016faster} to detect the boundary boxes of potential connectors, following similar approaches from previous work \cite{SundaresanGrannen-RSS-21, Shivakumar-RSS-22}. 

\begin{figure}[h]
\centering
\vspace{-0.25cm}
\includegraphics[width=\linewidth]{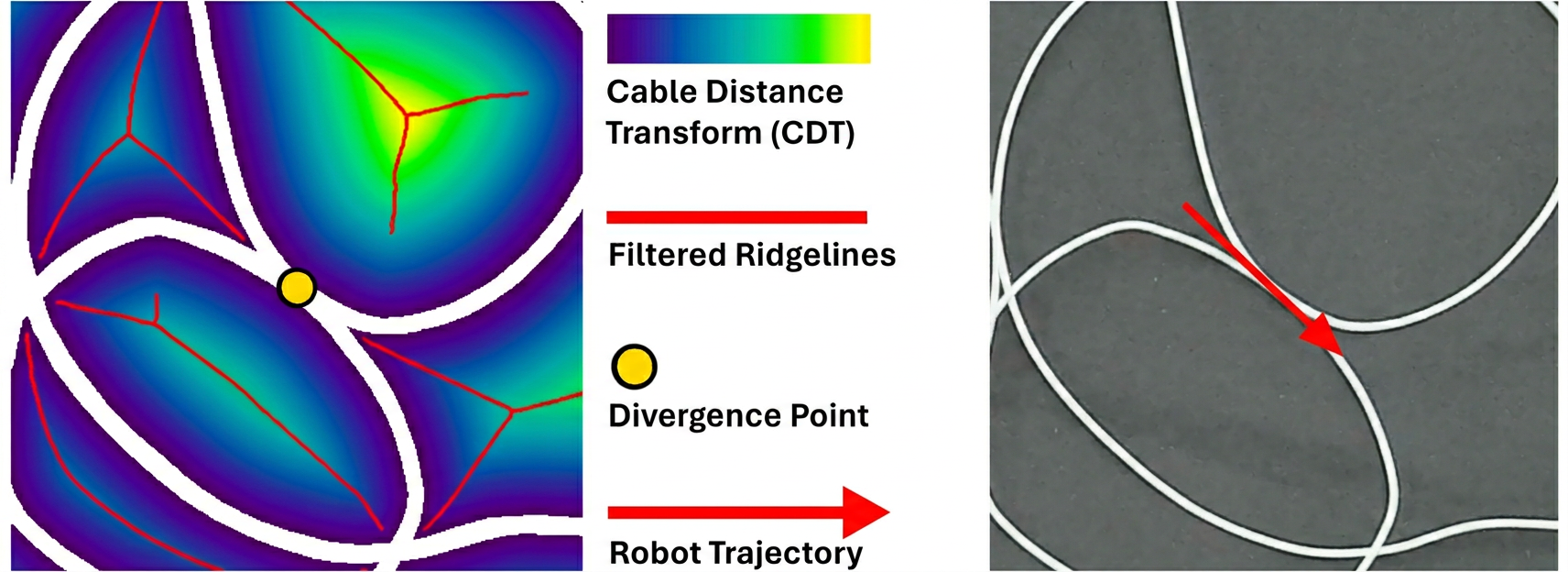} 
\caption{(Left): the \textbf{Cable Distance Transform (CDT)}, similar to a Voronoi diagram,  maps each pixel value to the Euclidean distance between the pixel and the nearest cable (white). Ridgelines (red) are from the CDT using a Hessian-based filtering algorithm and highlight potential paths for robot trajectory planning while minimizing unintended cable interaction. Divergence points (circled) indicate ambiguous intersections, resolved through targeted push primitives. (Right): a \textbf{Divergence Push} primitive starts along a ridgeline at one end and moves through the divergence point to attempt separation at tangential crossings.}
\label{fig:CDT}
\vspace{-0.25cm}
\end{figure}

\subsection{Bi-Directional Tracing}


To systematically identify ambiguous regions, \algabbr performs bi-directional tracing: it initiates a cable trace from every connector $e_i$ to produce two independent estimates for each cable. The system then identifies discrepancies where the two traces exhibit inconsistencies or fail to overlap to form a single continuous trajectory. When cables overlap in a tangential crossing, such as in Figure~\ref{fig:CDT}, visual ambiguities arise that make it difficult to distinguish individual paths.

\begin{figure}[h]
\centering
\vspace{0.05cm}
\includegraphics[width=1\linewidth]{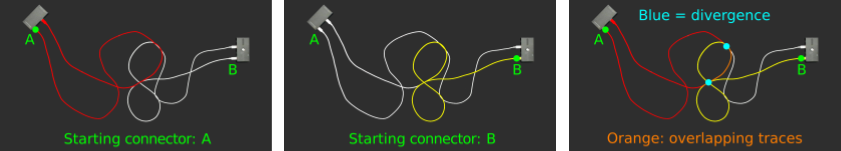} 
\caption{\algabbr's bi-directional tracing that identifies divergence points on 2 cables. (Left): cable traces initialized starting from 2 connectors of the same cable. (Right): overlapping cable traces are shown in orange; a divergence point in blue arises when the traces diverge into two non-overlapping paths.}
\label{fig:divergence}
\vspace{-0.25cm}
\end{figure}

\textbf{Divergence Points} occur when contradictory or incomplete connector assignments are present (many-to-one or one-to-none mappings), indicating that a cable trace $\hat{\theta}_{i}$ intersects another trace $\hat{\theta}_{j}$. Crossings at steep angles are visually unambiguous and are traced without contradictions, so they do not produce divergence points. Every divergence point therefore corresponds to a shallow-angle tangential crossing, a subset of crossings, by construction. To identify and disambiguate divergence points, at each timestep $t$, \algabbr executes in two phases:  

\textit{Phase 1: Reciprocal Consistency Verification}: \algabbr first identifies ambiguous cable paths by checking for reciprocal consistency—i.e., whether a trace from connector A terminates at connector B and a trace from connector B terminates at connector A (see Figure \ref{fig:divergence}). Traces that satisfy this bi-directional condition are marked as resolved. Trace pairs that fail (e.g., traces for one cable that diverge into two separate paths, or traces that terminate prematurely) are marked as unresolved and are passed to the next stage for divergence point detection.

\textit{Phase 2: Divergence Point Analysis}: For each unresolved trace, \algabbr searches for a candidate divergent trace pair using a hierarchical approach. It first attempts direct matching by searching for another trace whose terminal connector corresponds to the starting connector of the unresolved trace. If no direct match is found, \algabbr then searches from the resolved pairs from Phase 1 and chooses any pair that shares a connector with the unresolved trace. Lastly, \algabbr computes the overlap between the unresolved trace and all other traces and selects the pair with the highest overlap. The divergence point $d_{i}$ is the location where the two traces transition from overlapping to non-overlapping, as shown in Figure~\ref{fig:divergence}. Since all divergence points are shallow-angle overlaps by construction, they differ primarily in the number of cables involved. This number determines the interactive primitive best suited to resolve the ambiguity. We therefore classify divergence points based on local cable density $\rho(d_i)$, where $\rho(d_i)$ counts the number of distinct cables passing within a small radius $r$ around $d_i$. A \textit{cable cluster} occurs when $\rho(d_i)$ exceeds a threshold $\tau_c$, indicating high-density regions where cables visually merge, increasing the likelihood of erroneous connector assignments. In contrast, a \textit{tangential crossing} arises when $\rho(d_i) < \tau_c$, creating momentary overlaps without persistent occlusions or entanglements. The thresholds $r$ and $\tau_c$ are manually defined based on empirical observations of visually ambiguous scenarios where distinguishing between overlapping and non-overlapping traces is challenging. Formally, we define the following:

\vspace{-0.45cm}
\begin{equation}
    \mathcal{K}= \{ d_{i} \mid \rho(d_{i}) \geq \tau_c \}, \quad
    \mathcal{T}= \{ d_{i} \mid \rho(d_{i}) < \tau_c \}
\end{equation}
\vspace{-0.45cm}

\noindent where $\mathcal{K}$ represents the set of cable clusters, and $\mathcal{T}$ represents the set of tangential crossings.




\subsection{Two Novel Interactive Perception Primitives}

\algabbr also introduces two new interactive perception primitives: 1) \textit{Cluster Dilation}, which can physically separate densely packed cables located near trace points in set $\mathcal{K}$, and 2) \textit{Divergence Push}, which attempts to disambiguate tangential crossing points in set $\mathcal{T}$.

For robot trajectory planning in a multi-cable environment, we introduce the Cable Distance Transform (CDT), illustrated by Figure~\ref{fig:CDT}. The CDT, related to the Voronoi diagram \cite{voronoi}, is an adaptation of the Euclidean Distance Transform \cite{bailey2004efficient}. Given a binary mask generated from cable traces, CDT is defined as $D_c(p) = \min_{q \in \mathcal{C}} \| p - q \|$, where $D_c(p)$ computes the shortest Euclidean distance from pixel $p$ to the nearest cable pixel $q \in \hat{\Theta}_t$. The CDT maps open areas within densely packed cable environments to guide divergence push operations.

The CDT identifies ridgelines of maximum local distance change, which occur where a pixel is equidistant from two or more cables. These ridgelines act as a medial axis, defining the `safest' paths for robot trajectories by maximizing the clearance from neighboring cables.

\vspace{0.5cm}
\begin{figure*}[h]
  \centering
  \vspace{0.2cm}
  \renewcommand{\arraystretch}{0.7}
  \setlength{\tabcolsep}{3pt}
  \begin{tabular}{@{\hspace{3pt}} c @{\hspace{3pt}} c @{\hspace{3pt}} c @{\hspace{3pt}} c @{\hspace{3pt}}}
    \textbf{Tier 1}  & \textbf{Tier 2} & \textbf{Tier 3} & \textbf{Tier 4} \\ 
    \includegraphics[width=0.23\linewidth]{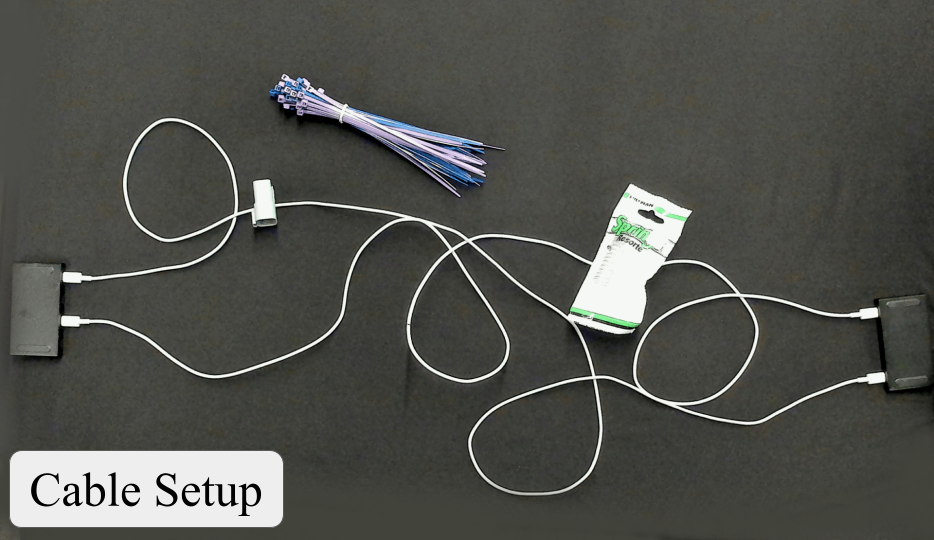} &
    \includegraphics[width=0.23\linewidth]{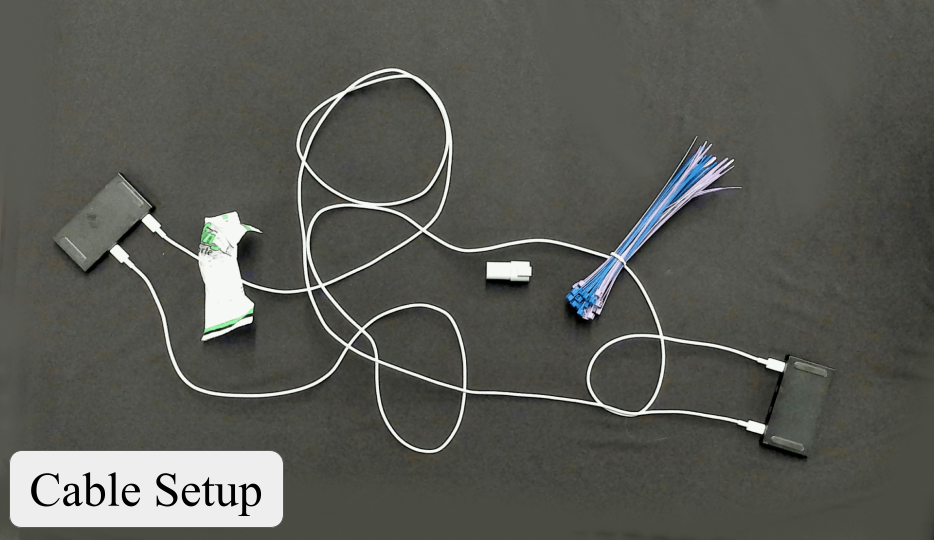} &
    \includegraphics[width=0.23\linewidth]{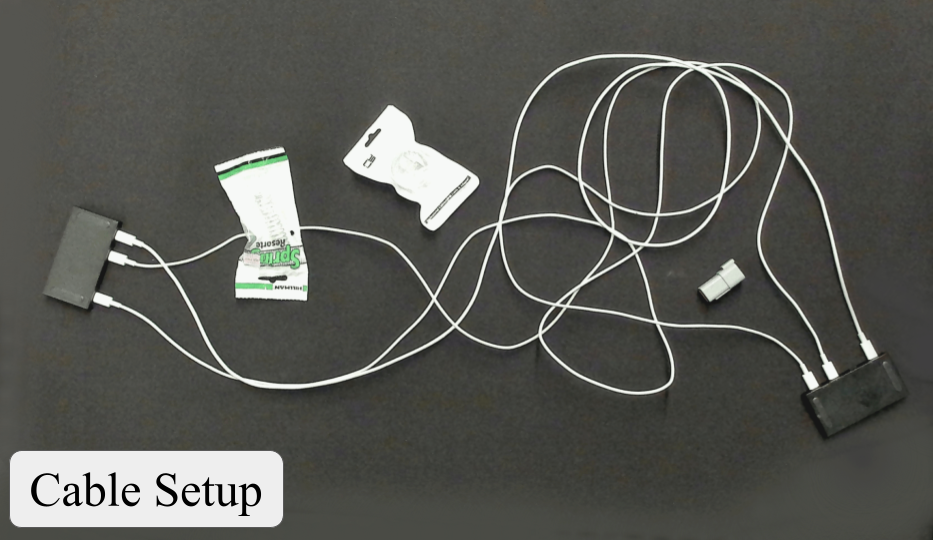} &
    \includegraphics[width=0.23\linewidth]{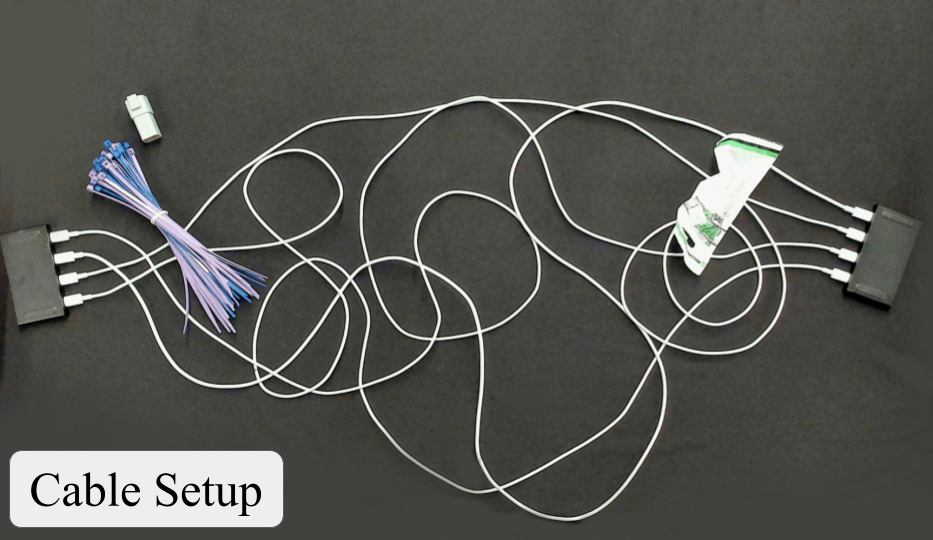} \\
    \includegraphics[width=0.23\linewidth]{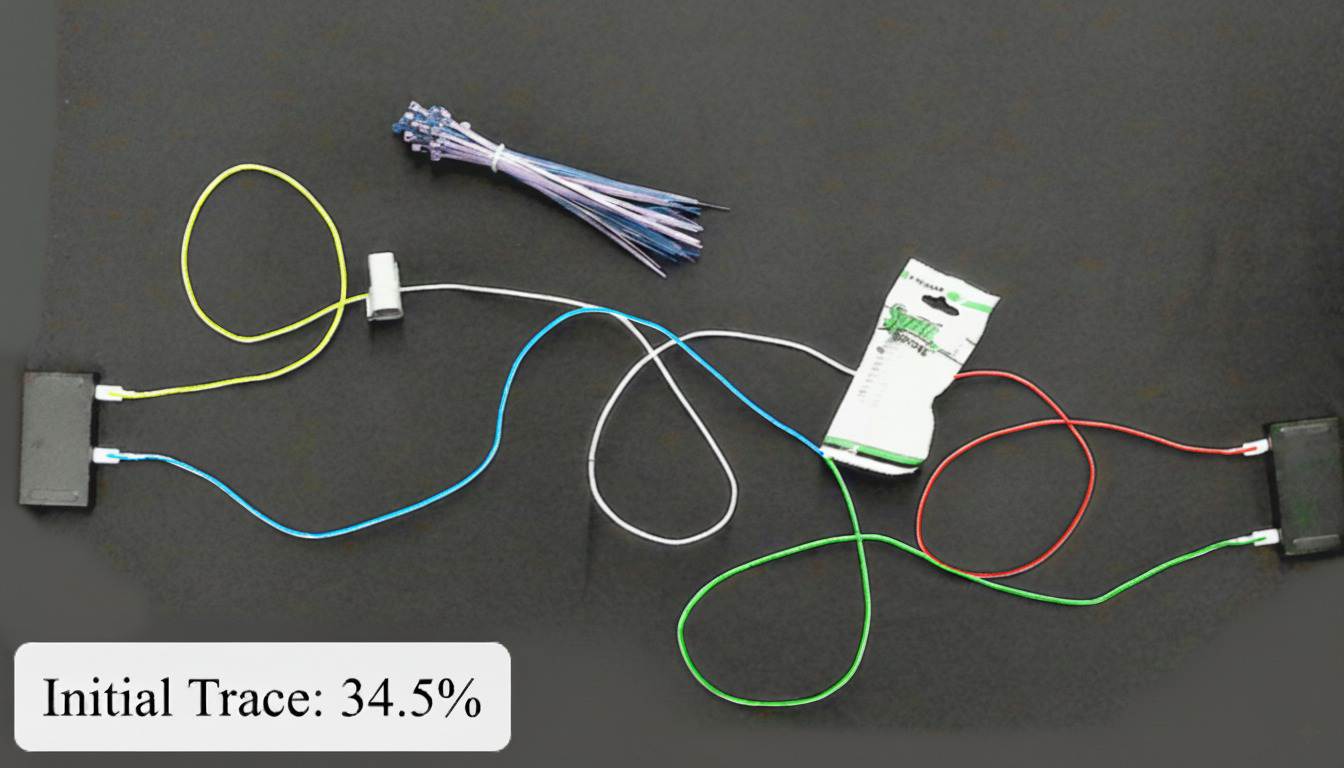} &
    \includegraphics[width=0.23\linewidth]{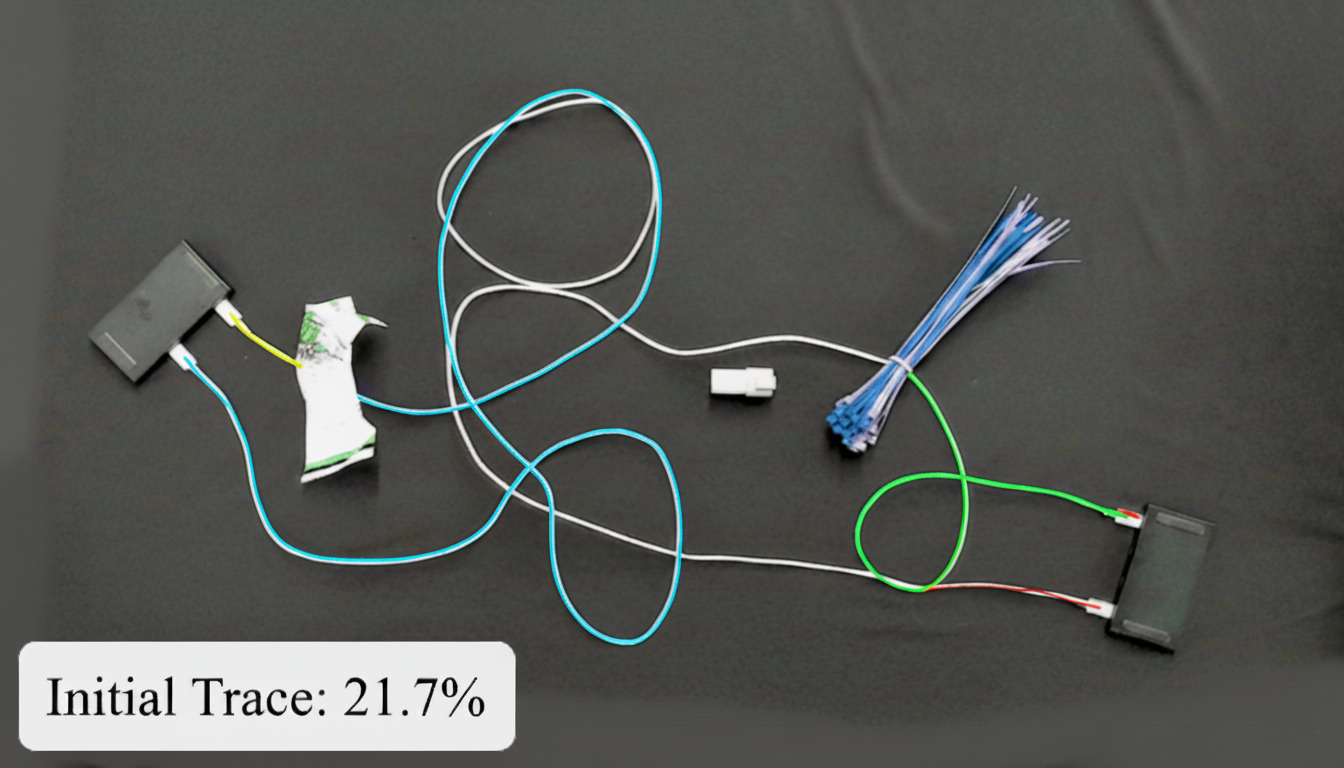} &
    \includegraphics[width=0.23\linewidth]{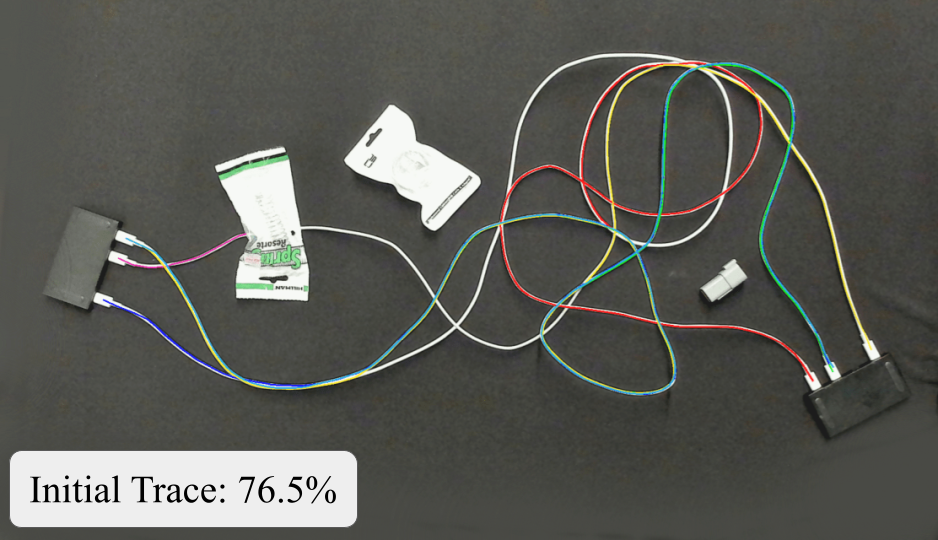} &
    \includegraphics[width=0.23\linewidth]{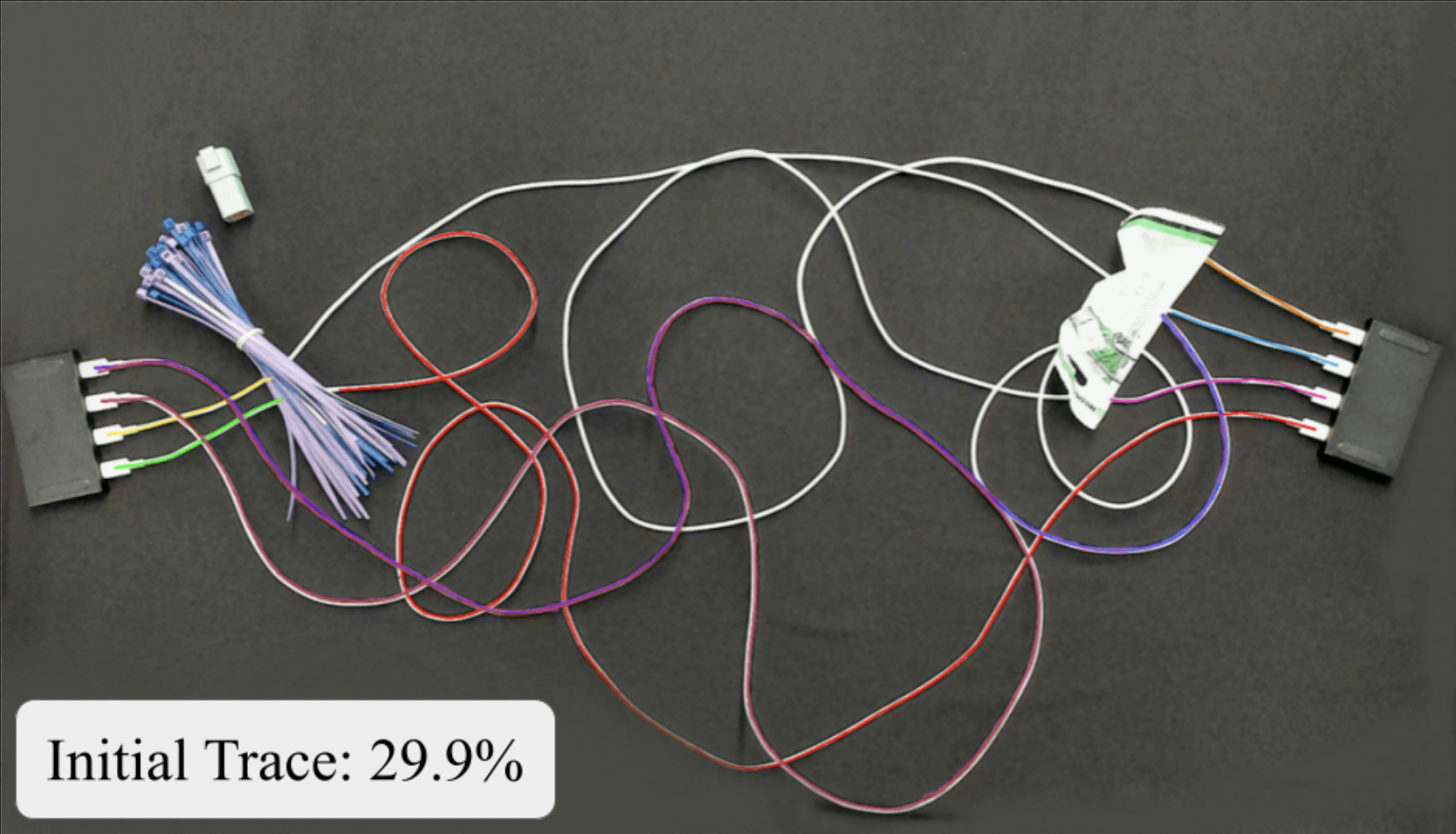} \\
    \includegraphics[width=0.23\linewidth]{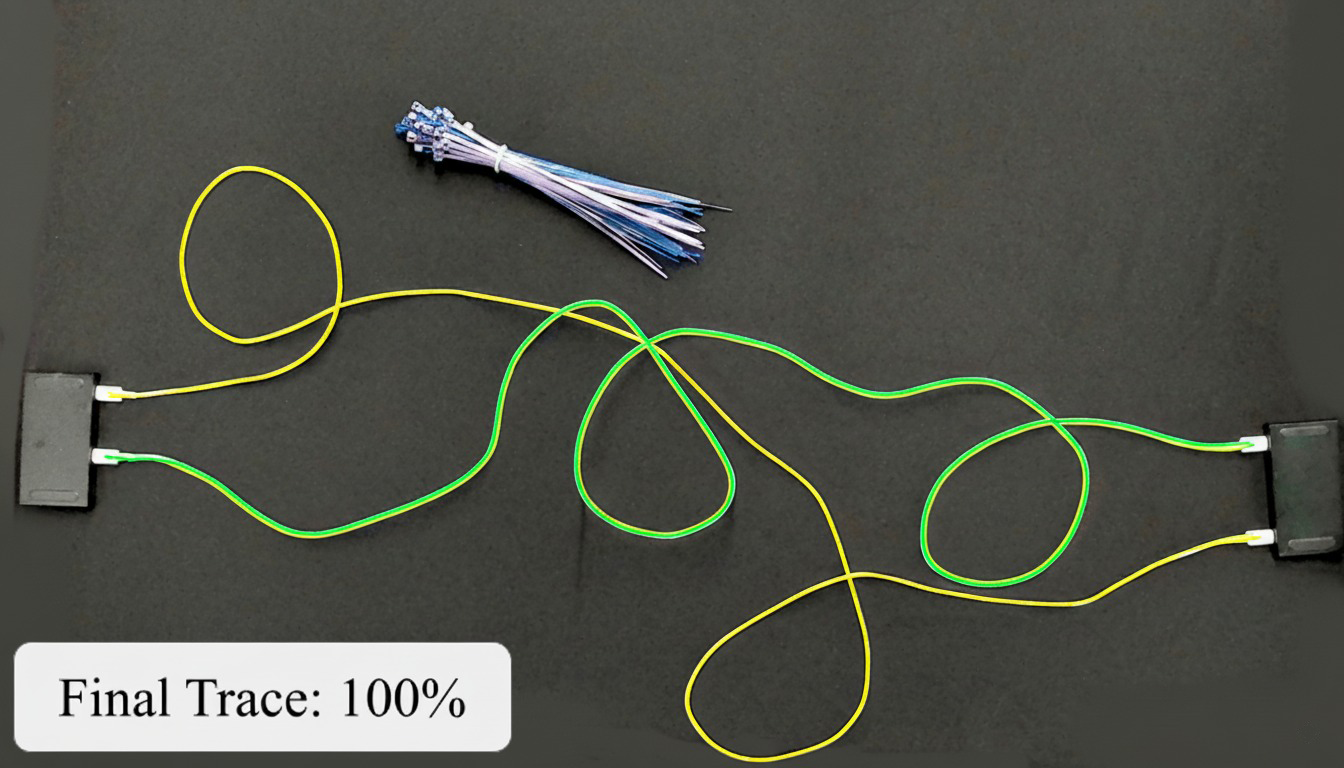} &
    \includegraphics[width=0.23\linewidth]{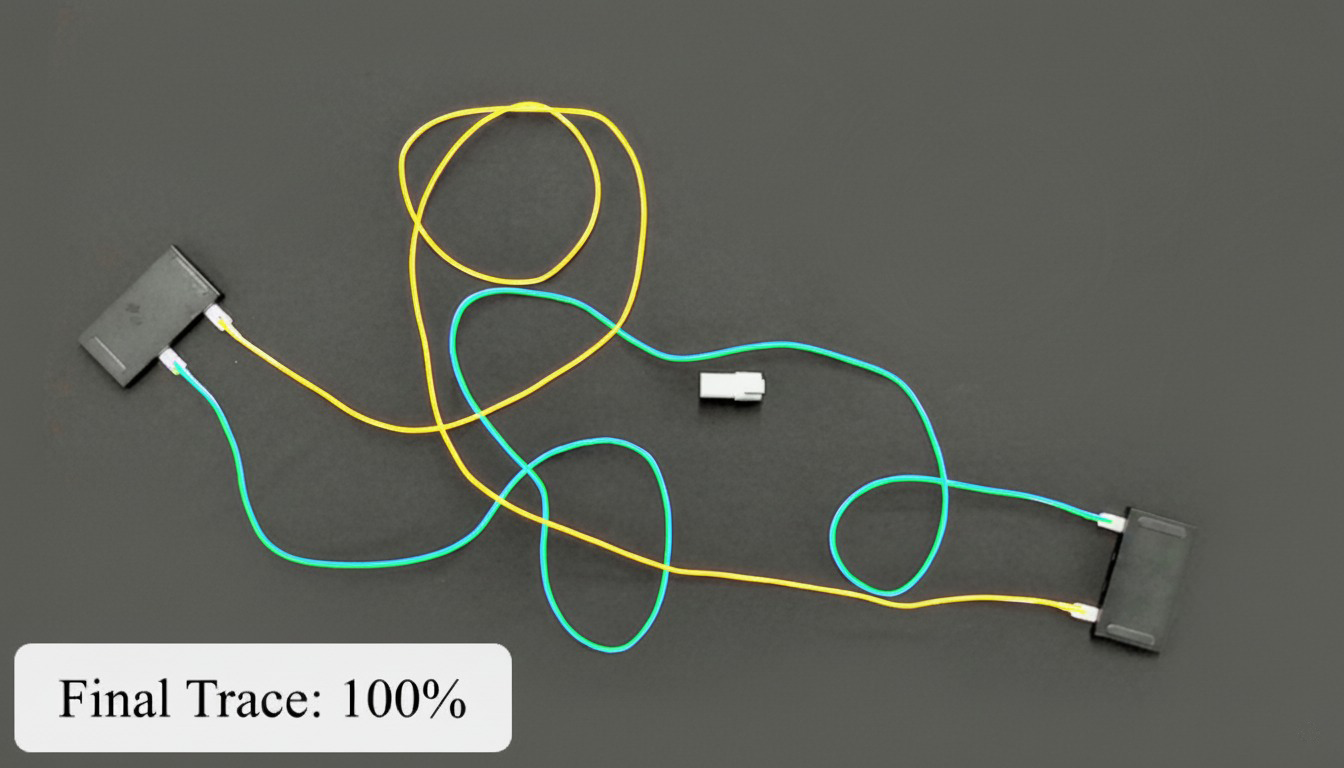} &
    \includegraphics[width=0.23\linewidth]{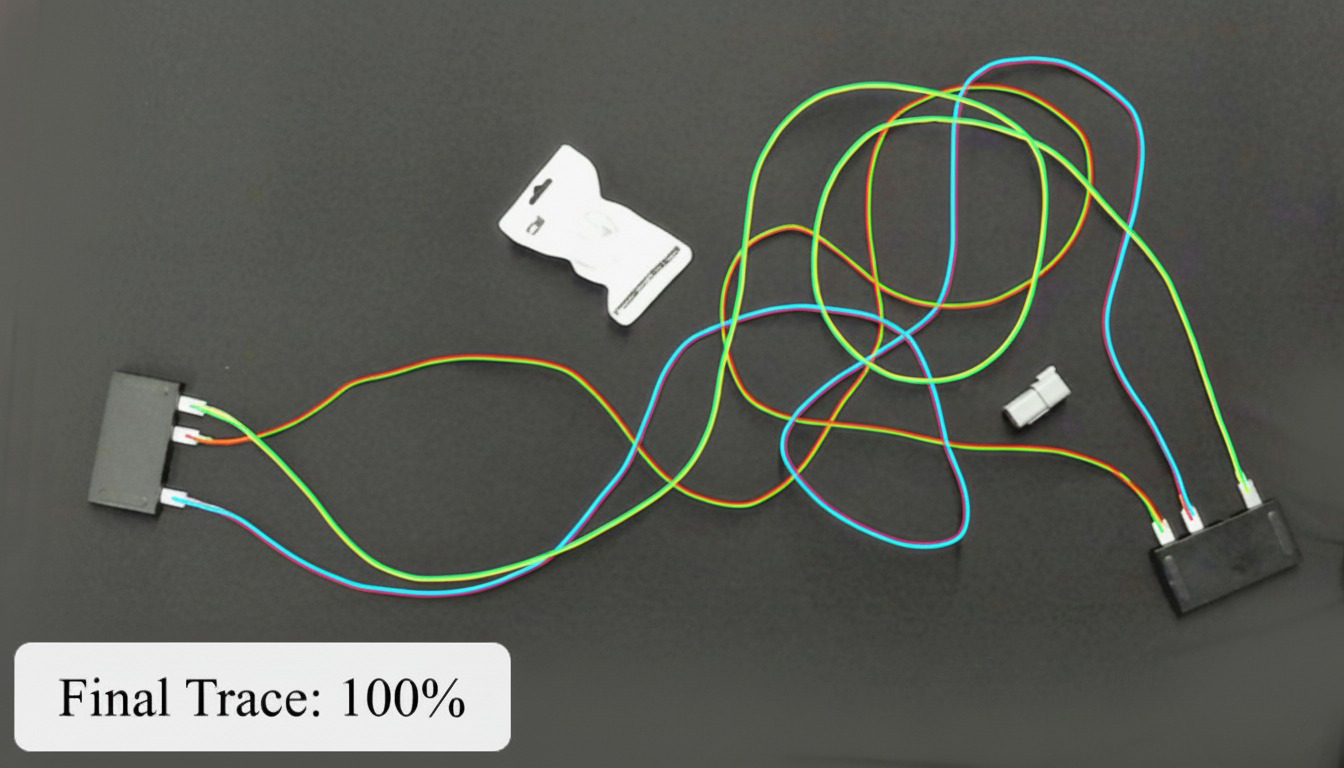} &
    \includegraphics[width=0.23\linewidth]{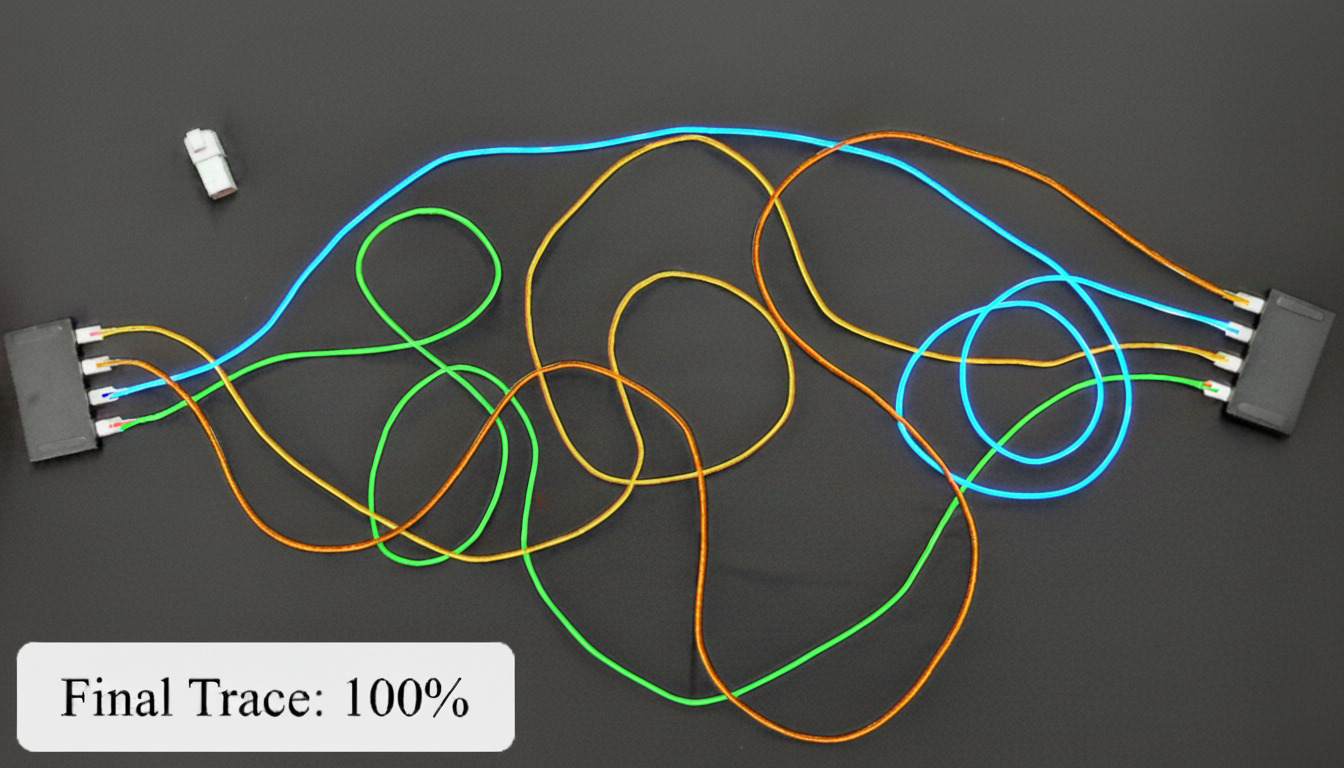} \\
  \end{tabular}
  \caption{Four examples; each column represents a different tier. The first row shows the initial scene with cluttered cables and external objects, the second row displays the initial predicted traces with occlusions, the third row shows the completed traces after object decluttering and interactive perception primitives.}
  \label{fig:tier_system}
  \vspace{-0.25cm}
\end{figure*}
\vspace{-0.45cm}
\textbf{Cluster Dilation:} To separate clusters, \algabbr identifies open regions near each $d_i \in \mathcal{K}$. Open regions $\mathcal{O}$ are defined as non-cable pixels whose CDT satisfies $D_c(p) > \tau_o \,$, where $\tau_o$ is a separation threshold. Regions are filtered by area constraints $\mathcal\alpha_{min} \leq A(\mathcal{O}) \leq \alpha_{max} $ to ensure feasible gripper insertion. The system selects the largest open region and moves a closed gripper to pixel $p^*=\operatorname{argmax}_{p \in R} D_c(p)$. The gripper fully opens and applies two 180\textdegree\ rotations in opposite directions in an attempt to separate the cables.



\textbf{Divergence Push:}
TRACE performs ridge detection on the CDT using the Frangi vesselness filter~\cite{10.1007/BFb0056195} shown in Figure~\ref{fig:CDT} as red lines. This filter analyzes the second-order local structure of the CDT by computing the eigenvalues of the Hessian matrix, which highlights the structures corresponding to ridges. The resulting ridge set  $\mathcal{R}$  is defined as: $\mathcal{R} = \{ p \mid V(p) > \tau_r \}$, where $\tau_r$ is a threshold to distinguish significant ridges from the background. The vesselness measure $V(p)$ quantifies the likelihood that a pixel $p$ belongs to a ridge-like structure by evaluating the ratio and magnitude of the eigenvalues. High vesselness values correspond to regions where the Hessian eigenvalues indicate elongated linear features, which naturally align with the medial axes between cables. 

In practice, following these ridges often reduces unintended cable interaction. To determine the push direction, TRACE computes the tangential lines of the crossing cables. The push interaction starting point is then initialized along the ridge set $\mathcal{R}$, constrained by the two tangential lines and a predefined keep-out distance $d_{\text{safe}}$. We set $d_{\text{safe}} = 12$ pixels, a value derived from the physical width of the robot gripper when projected into the image pixel space. The push action is executed along this ridge: beginning at the push interaction starting point, sliding the end-effector through the divergence point, and continuing for an equal distance beyond the divergence point.

\begin{table}[h]
\centering
\caption{Tiers of complexity for evaluation} 
\label{tab:tier_system}
\begin{tabular}{lcccc}
\toprule
\textbf{} & \textbf{Tier 1} & \textbf{Tier 2} & \textbf{Tier 3} & \textbf{Tier 4} \\
\midrule
\textbf{\# Cables} & 2 & 2 & 3 & 4 \\
\midrule
\textbf{Avg \# Crossings} & 12 & 15 & 30 & 40\\
\textbf{\# Tangential Crossings} & 2 & 3 & 3-4 & 4-5\\
\textbf{\# Foreground Objects} & 3-4 & 3-4 & 3-4 & 3-4\\
\bottomrule
\end{tabular}
\vspace{-0.25cm}
\end{table}

\subsection{Pipeline and Termination} We formalize the complete pipeline of \algabbr. At each timestep $t$, the system receives an image observation $\ve{I_t}$ and detects external objects and cable connectors $\mathcal{E}(\ve{I_t}) = \{ e_1, e_2, \dots, e_n \}$ using a trained connector detection model. Each detected connector $e_i$ produces an estimated trajectory $\hat{\theta}_i$.


If any external objects actively occlude the cable trace, the system applies an object removal action. It then identifies divergence points ($\mathcal{D}_t$), classifying them as either tangential crossings ($\mathcal{T}_t$) or high-density cable clusters ($\mathcal{K}_t$). Then, the system applies the associated interactive perception action $\ve{a_t}$ to update the estimated complete cable state $\hat{\Theta}_{t+1}$, improving visibility and refining trajectory estimates $\hat{\theta}_{i, t+1}$. This process iterates until the termination condition $T_{max}=10$ timesteps is reached or until all ambiguities are resolved such that the reconstructed cable set $\hat{\Theta} = \bigcup_{i=1}^{n} \hat{\theta}_i$ is topologically consistent, meaning each cable is reconstructed as a single continuous curve that maintains correct connectivity through crossings without identity errors. 
\section{Experiments}

\textbf{Hardware:}
The experimental setup consists of a bimanual (to increase the effective workspace reachability) ABB YuMi robot with motion planning trajectories solved by Jacobi Motion ~\cite{jacobi2024motion}, and an overhead Logitech BRIO RGB camera positioned 1 m above the workspace. The robot operates over a planar workspace, where 2-4 standard white ~1.8 m USB-C to USB-C cables are randomly arranged. The connectors of each cable are plugged into two 8x5 cm black USB hubs on opposite sides of the workspace. The overhead camera captures 4K resolution RGB images of the cable environment. 


\begin{table}[t]
\centering
\caption{Average number of interactive perception primitives used per
trial (over 60 trials), reported as count (\% of total). Cluster dilation
is used more as scene context increases.}
\setlength{\tabcolsep}{4pt}
\footnotesize
\begin{tabular}{lrrrr}
\toprule
\textbf{Primitive} & \textbf{Tier 1} & \textbf{Tier 2} & \textbf{Tier 3} & \textbf{Tier 4} \\
\midrule
Bimanual Decl. & 1.19 (36.5) & 1.53 (34.8) & 1.33 (19.2) & 1.27 (16.1) \\
Cluster Dilation & 0.19 (5.8) & 0.53 (12.1) & 1.00 (14.4) & 2.80 (35.6) \\
Divergence Push & 1.88 (57.6) & 2.33 (53.1) & 4.60 (66.4) & 3.80 (48.3) \\
\midrule
Total & 3.25 (100) & 4.40 (100) & 6.93 (100) & 7.87 (100) \\
\bottomrule
\label{tab:types_of_moves}
\end{tabular}
\vspace{-1.0cm}
\end{table}

We evaluate the performance of \algabbr with varying numbers of cables and foreground objects. The evaluation follows a tiered framework that is described in Table \ref{tab:tier_system} and in Figure~\ref{fig:tier_system}.


\subsection{Background Clutter}

In real world environments such as data centers, cables are rarely placed on uniform or texture-less surfaces. Background clutter such as distractor objects can introduce spurious edges or color variation that interfere with cable tracing and divergence detection. To evaluate robustness of background variability, we conduct experiments across multiple conditions with various levels of background visual complexity. Figure \ref{fig:background_clutter} shows an example TRACE pipeline run on a scenario with white cables and a cluttered background.

Preliminary results indicate that while performance degrades under cluttered backgrounds, the system continues to have stable divergence detection and interactive perception primitives. In these experiments, we additionally used a different set of cables with increased stiffness and spring-back behavior compared to the main evaluation, increasing trace difficulty. Table \ref{tab:background_clutter} presents the average percent of cable correctly traced on configurations with background clutter across different complexity levels. 

\begin{table}[h]
\vspace{-0.15cm}
\centering
\caption{Average percent of cable length correctly traced (20 trials).}
\label{tab:background_clutter}
\begin{tabular}{lcccc}
\toprule
 & \textbf{Tier 1} & \textbf{Tier 2} & \textbf{Tier 3} & \textbf{Tier 4} \\
\midrule
\text{TRACE (No Clutter)} 
& 99.1\% & 91.2\% & 94.2\% & 89.4\% \\
\text{TRACE (Background Clutter)} 
& 78.7\% & 77.3\% & 66.2\% & 65.3\% \\
\bottomrule
\end{tabular}
\vspace{-0.40cm}
\end{table}

\begin{figure*}[h]
\centering
\vspace{0.25cm}
\includegraphics[width=\linewidth]{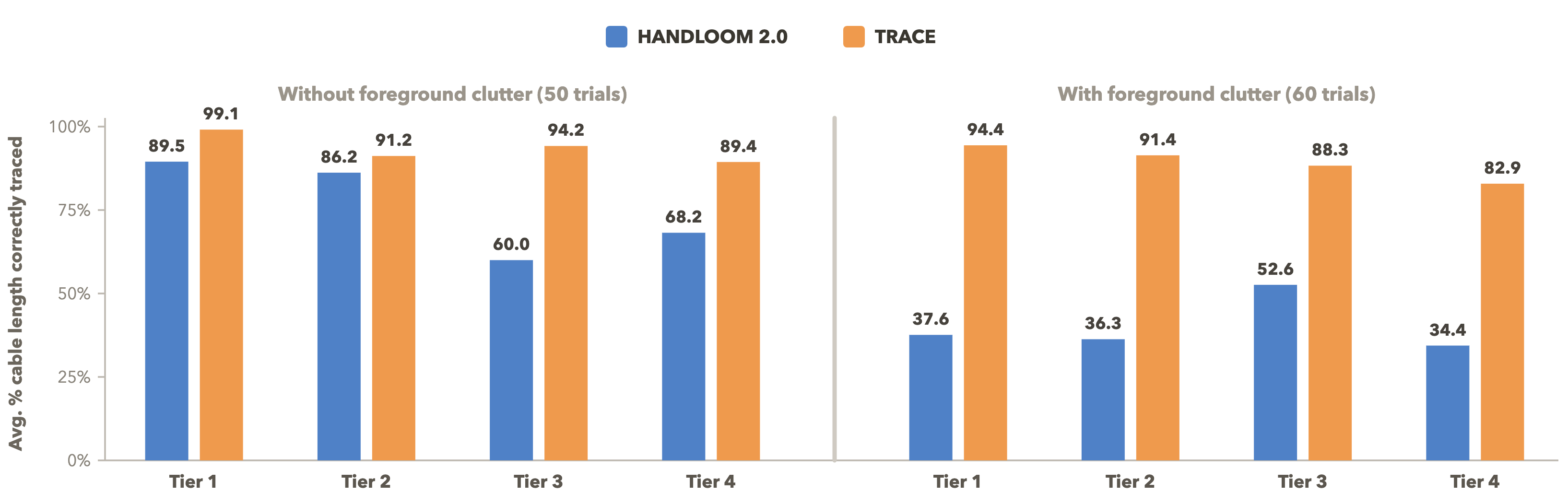} 
\caption{\small (Left) Without foreground clutter---Average percent of cable length correctly traced comparing \algabbr and HANDLOOM 2.0 for 50 physical experiments. Tracing results for HANDLOOM 2.0 were sourced from MANIP \cite{yu2024manip}. (Right) With foreground clutter---Average percent of cable length correctly traced using \algabbr with bimanual decluttering for 60 physical experiments. Vertical bars represent one standard deviation of the mean.}
\label{fig:results}
\vspace{-0.5cm}
\end{figure*}

\begin{figure}[h]
\centering
\includegraphics[width=1\linewidth]{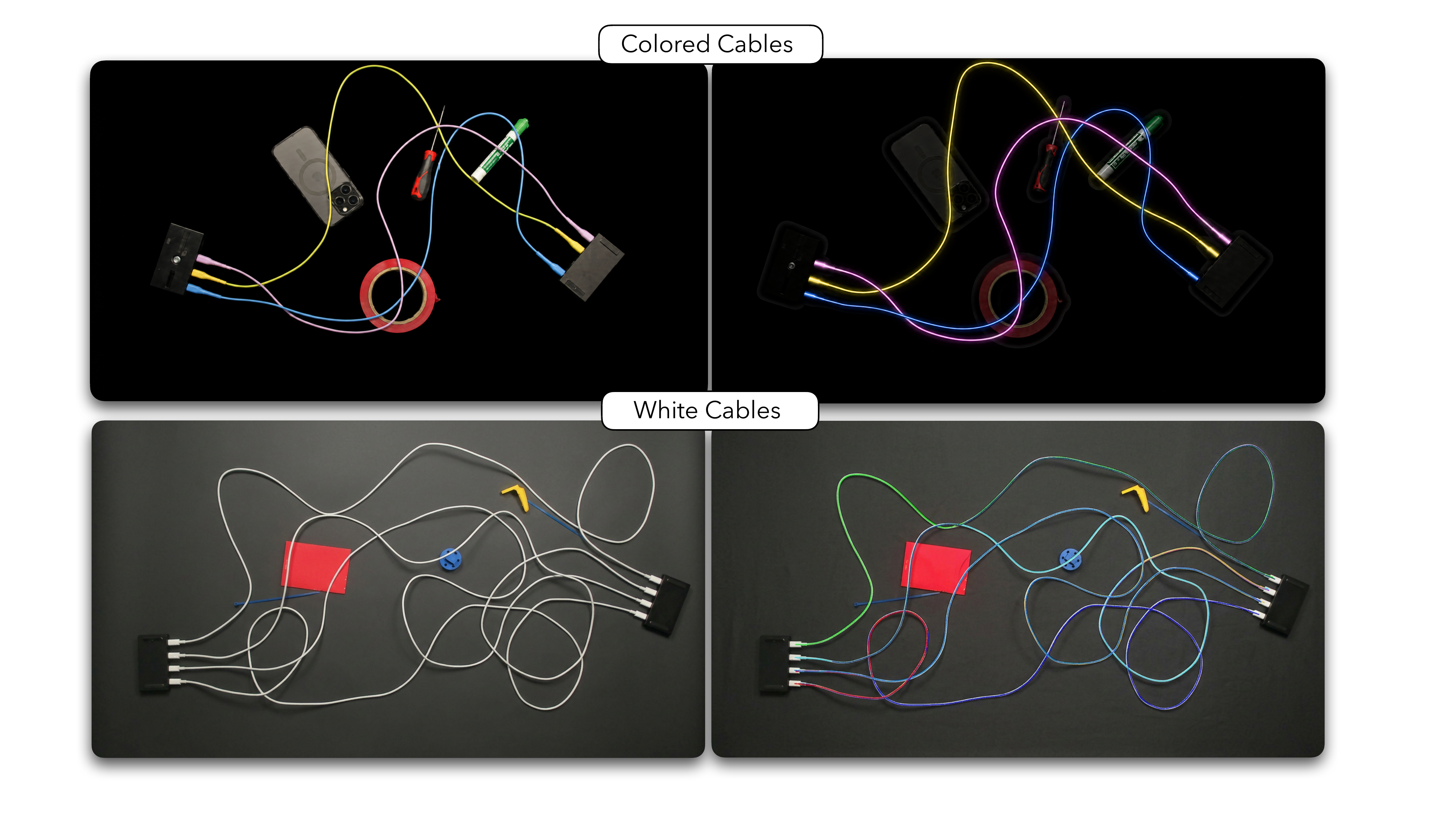} 
\caption{Colored cables are relatively easy to trace in contrast to monochrome cables: \algabbr execution with background clutter, with colored cables (Top) and white cables (Bottom). (Top Left) Initial scene with colored cables and background clutter. (Top Right) Color-thresholded segmentation masks correctly trace 100\% of cables. (Bottom Left) The initial configuration of white cables placed on a visually complex tabletop with distractor textures and objects. (Bottom Right) The reconstructed cable traces after bi-directional tracing and the application of interactive perception primitives, demonstrating robustness to spurious edges and background variability. }
\label{fig:background_clutter}
\vspace{-0.65cm}
\end{figure}

\subsection{Colored Cables}

\algabbr is designed specifically for the challenging case of monochrome cable configurations, where individual cables are visually indistinguishable. We conduct a case study on colored cable configurations as shown in Figure \ref{fig:background_clutter}. Naive HSV color thresholding around each cable connector reliably separates individual cable instances, significantly reducing ambiguity at crossings and yielding high performance on cable connector matching and tracing.



\subsection{Multi-Cable Tracing Evaluation}

We first evaluate the \algabbr framework in an ablation study evaluating against HANDLOOM 2.0, which does not use bi-directional tracing. We conducted 50 physical experiment trials (without foreground objects) distributed across the tiers, with 13 trials in Tiers 1-3 and 11 trials in Tier 4. For each tier, we report the correctly traced cable length relative to the total known cable length in the scene, which is empirically measured. Figure \ref{fig:results} presents the average percent of cable length correctly traced comparing both models across different complexity levels. Trace percentages for HANDLOOM 2.0 were sourced from experiments in MANIP \cite{yu2024manip} (independent from the 50 trials for \algabbr). Results suggest that \algabbr significantly outperforms HANDLOOM 2.0, achieving comparable correct trace performance Tier 1 and Tier 2 trials while demonstrating a substantial improvement in Tier 3 and Tier 4 trials.

\subsection{With Foreground Object Clutter}

We conducted 60 physical experiment trials with clutter objects on top of cables, evenly distributed across the 4 tiers. In 32 out of 60 trials, \algabbr correctly traced 100\% of all cables. Figure \ref{fig:results} presents the average percent of cable length correctly traced with foreground objects. While performance slightly decreases compared to the clutter-free setting, \algabbr averages 94.4\% in Tier 1 to 82.9\% in Tier 4. Additionally, it significantly outperforms HANDLOOM 2.0 on identical scenes by an average of 77.0\% improvement across all tiers. Table~\ref{tab:types_of_moves} illustrates the average number of interactive perception primitives across different tiers. In simpler scenes (Tiers 1–2), Divergence Push is the dominant action, while in more complex configurations (Tiers 3–4), Cluster Dilation becomes more prevalent.

\vspace{0.25cm}
\begin{figure}[h]
\vspace{0.15cm}
\centering
\includegraphics[width=1\linewidth]{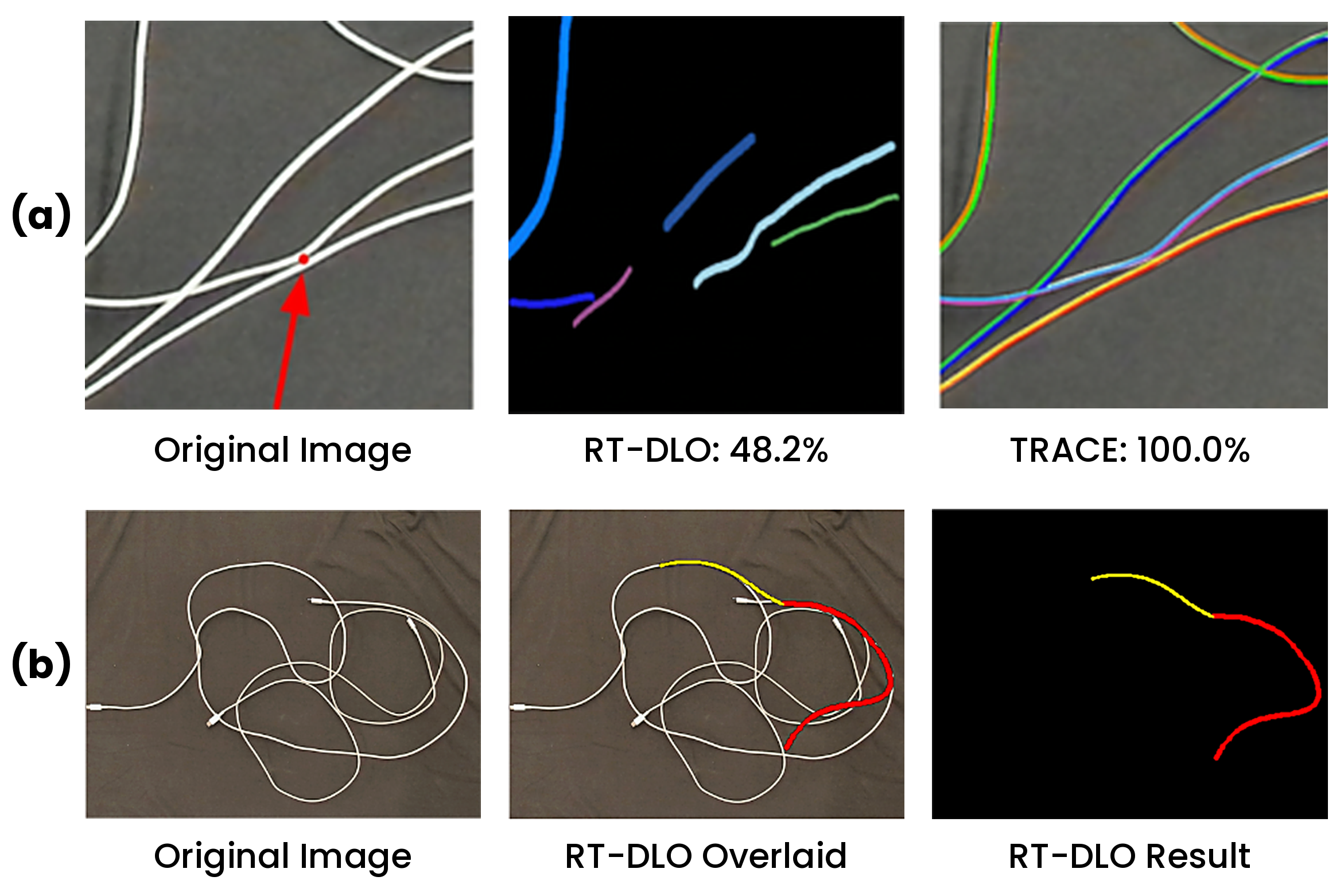}
\caption{Baseline method RT-DLO performs poorly even on non-cluttered backgrounds, where cables
are visually unambiguous. \textbf{(a)} On a cropped region, RT-DLO correctly
traces only 48.2\% of the total cable length while \algabbr{} traces 100.0\%;
the red arrow marks the divergence point at which RT-DLO's trace fails.
\textbf{(b)} On a full cable scene, RT-DLO recovers only a fraction of the
cable length: (left) original image, (center) RT-DLO trace overlaid on the
original, and (right) the isolated RT-DLO result.}
\label{fig:rtdlo-comparison}
\vspace{-0.75cm}
\end{figure}

\subsection{Comparison with RT-DLO}

We compare \algabbr to RT-DLO \cite{RTDLO} from 2023, the most comparable tracing system evaluated on cables of similar thickness and color. Although RT-DLO focuses on tracing short, multi-colored cables in the presence of background clutter, it has not been evaluated on three or more longer cables, higher crossing counts, or high-density clusters. 

To study these differences and assess the effectiveness of \algabbr, we use the divergence points identified by \algabbr as regions of ambiguity and select 10 random samples among these across all 4 tiers. For each trial, we locate these divergence points in the ``before'' image (the raw initial setup) and crop a local 600x600 pixel region around them. We then run RT-DLO on this cropped segment and calculate the average proportion of all cables correctly traced. After \algabbr completes all interactive perception primitives, we then run RT-DLO again on the corresponding ``after" image, and calculate the final trace percentage. These initial and final results are then compared with those produced by \algabbr. Figure \ref{fig:rtdlo-comparison}(a) shows a comparison between
percent of cable length correctly traced on an image crop
evaluated by both systems on a non-cluttered background.


After \algabbr interactive perception primitives, RT-DLO's accuracy improves, as shown in Table~\ref{tab:rtdlo_baseline}. However, the \algabbr algorithm significantly outperforms RT-DLO on small crops. We also explored RT-DLO's performance on scenes with long cables. As illustrated in Figure~\ref{fig:rtdlo-comparison}(b), when applied to a full scene, \algabbr outperforms RT-DLO. 

\begin{table}[h]
\centering
\caption{Average percentage of cable length correctly traced in small image crops, comparing \algabbr and RT-DLO over 10 trials without background clutter.}
\label{tab:rtdlo_baseline}
\begin{tabular}{lcccc}
\toprule
\textbf{Method} & \textbf{Computation Time} & \textbf{Initial} & \textbf{Final} & \textbf{Improvement}\\
\midrule
RT-DLO & 0.05s & 57.1\% & 76.0\% & 18.9\%\\
\algabbr & 0.40s & 68.3\% & \textbf{97.5\%} & 29.2\%\\
\cmidrule{1-5}
\textbf{Increase} & & 19.6\% & 28.3\% \\
\bottomrule
\end{tabular}
\vspace{-0.25cm}
\end{table}

\subsection{Comparison Against State-of-the-Art VLMs}
As modern general-purpose vision-language models (VLMs) demonstrate strong performance on zero-shot visual reasoning tasks, we evaluate Nano Banana Pro and ChatGPT 5.2 to assess whether high-level semantic reasoning alone is sufficient for cable tracing scenarios. We randomly selected five physical experiments each from Tier 1 through Tier 4 from the pool of Figure \ref{fig:results}. We provide the ``before'' RGB image (the raw initial setup) and the ``after'' RGB image after TRACE completes all interactive perception primitives for a total of 40 images for each VLM. The models were instructed to output predicted cable traces by recoloring each cable in the image with a unique color with the following prompt:
\begin{tightquote}
``Analyze the provided image and trace each distinct cable as a continuous object from one visible connector to another. For this task, the start point for tracing each cable is defined as its connector on the black rectangular device on the left side of the image. The end point for each cable is its connector on the device on the right side of the image. Carefully trace each cable from its start point to its end point, even if they cross or overlap. Assign each entire cable a unique, clearly distinguishable color (e.g., bright red, blue, green, yellow, purple, orange). Produce a modified version of the original image where each entire cable is recolored consistently with its assigned unique color from its start point to its end point. Do not color any non-cable objects. Do not merge cables that only visually cross, ensuring that every visible cable segment is fully covered by exactly one unique color.''
\end{tightquote}
We evaluated performance by computing the average proportion of total cable length correctly traced across all cables in each scene. Trace length was measured relative to the empirically-known ground truth cable topology. If the model hallucinated a cable, the prediction was assigned a score of 0. Results are shown in Table \ref{tab:nano}.

\begin{figure}[h]
\centering
\vspace{-0.25cm}
\includegraphics[width=1\linewidth]{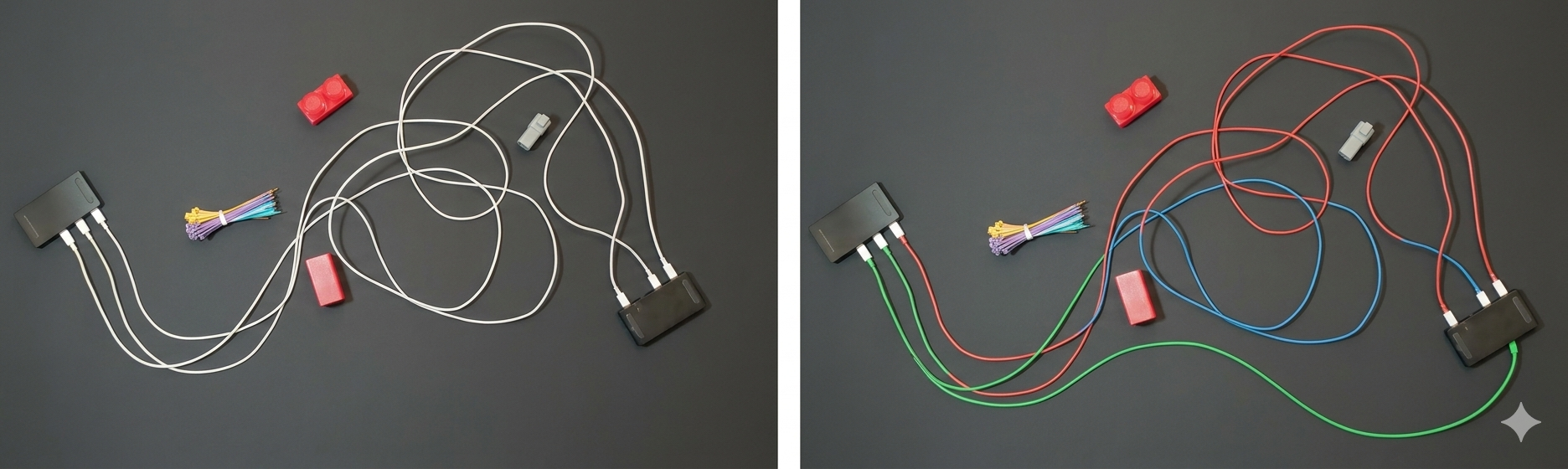}
\caption{Nano Banana Pro outputs on complex cable scenes, showing fragmented and topologically inconsistent traces compared to TRACE. (Left) Initial scene. (Right) Nano Banana Pro output. Note the hallucinated green cable appearing from the bottom of the right USB hub.}
\label{fig:nano}
\vspace{-0.25cm}
\end{figure}
An example trial of Nano Banana Pro is shown in Figure \ref{fig:nano}. While both Nano Banana Pro and ChatGPT 5.2 demonstrate strong zero-shot segmentation capability, both do not reliably trace cables through crossings and high-density clutter, reflecting the absence of explicit connectivity consistency. Both models frequently hallucinate cable segments or lose portions of valid cable traces, demonstrating that high-level semantic reasoning is insufficient for consistent tracing of monochrome cables.

\begin{table}[h]
\centering
\caption{Average percentage of cable length correctly trace in full-scene images with foreground clutter, comparing TRACE to Nano Banana Pro and ChatGPT 5.2. Initial and Final correspond to images taken before and after the interactive perception primitives.}
\label{tab:nano}
\begin{tabular}{lcc}
\toprule
\textbf{Method} & \textbf{Initial} & \textbf{Final}\\
\midrule
Nano Banana Pro & 37.5\% & 34.8\%\\
ChatGPT 5.2 & 19.5\% & 26.8\%\\
\textbf{\algabbr} & \textbf{40.2\%} & \textbf{89.3\%}\\
\bottomrule
\end{tabular}
\vspace{-0.55cm}
\end{table}

\subsection{Experiments with 6 and 8 Cables}

\begin{figure}[h]
\centering
\vspace{-0.25cm}
\includegraphics[width=1\linewidth]{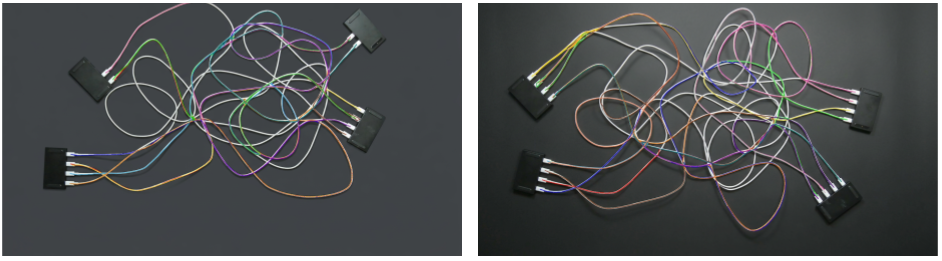} 
\caption{\algabbr on 2 scenes with increased cable count (6 cables, left; 8 cables, right). Cables densely cover nearly the entire workspace, leaving minimal visible background and creating highly occluded scenes.}
\label{fig:6-8}
\vspace{-0.25cm}
\end{figure}

As illustrated in Figure \ref{fig:6-8}, we also perform experiments with 6 and 8 cables using 4 USB hubs. These examples significantly increase cable density, include up to 120 crossings and 16 tangential crossings, and reduce the reachable space for the robot gripper to perform interactive perception primitives, all of which degrade \algabbr's performance. Although all 12 and 16 cable connectors, respectively, were identified by \algabbr, the pipeline was not able to resolve divergence points and was only able to correctly trace 30\% of total cable lengths. 

\section{Limitations and Future Work}

As noted in the previous section, \algabbr's performance degrades with increased cable density. One way to address this may be to segment the scene into smaller rectangular regions, initiate cable tracing from edge points where cables exit each region, and stitch together these traces by matching cable exit points from each region with those of adjacent regions. We will also consider narrower gripper jaws that can be more precisely inserted into dense cable regions for interactive perception.


\section{Conclusion}

We present TRACE, a framework for tracing multiple monochrome cables in cluttered environments from monocular RGB images alone.
TRACE combines bi-directional tracing, which identifies regions of uncertainty and gaps in connectivity, with two interactive perception primitives, Divergence Push and Cluster Dilation, that actively resolve crossings and occlusions. Across 110 physical experiments, TRACE increases the percentage of cable length correctly traced from $\sim$60\% with the strongest prior method, HANDLOOM 2.0, to $\sim$90\% in complex multi-cable configurations, substantially outperforming RT-DLO and recent VLMs. These results suggest that combining global connectivity analysis with targeted interactive perception is a promising direction for deformable linear object state estimation.

\renewcommand*{\bibfont}{\footnotesize}
\printbibliography
\end{document}